\documentclass[conference]{IEEEtran}
\IEEEoverridecommandlockouts
\usepackage{cite}
\usepackage{amsmath,amssymb,amsfonts}
\usepackage{algorithmic}
\usepackage{graphicx}
\usepackage{textcomp}
\usepackage{xcolor}
\usepackage{bbm}
\usepackage{multirow}
\usepackage{array}
\usepackage{url}
\usepackage{booktabs}

\def\BibTeX{{\rm B\kern-.05em{\sc i\kern-.025em b}\kern-.08em
    T\kern-.1667em\lower.7ex\hbox{E}\kern-.125emX}}
\begin{document}

\title{Cross-Field Transformer for Diabetic Retinopathy Grading on Two-field Fundus Images\\
\thanks{This work was supported by National Natural Science Foundation of China (No. 62172101), the Science and Technology Commission of Shanghai Municipality (No. 22511106000; No. 21511104506; No. 22692117100). }
}


\author{
\IEEEauthorblockN{
Junlin Hou\textsuperscript{1},
Jilan Xu\textsuperscript{1},
Fan Xiao\textsuperscript{2},
Rui-Wei Zhao\textsuperscript{2},
Yuejie Zhang\textsuperscript{1,*},\\
Haidong Zou\textsuperscript{3,4},
Lina Lu\textsuperscript{4},
Wenwen Xue\textsuperscript{4},
Rui Feng\textsuperscript{1,2,*}
}
\IEEEauthorblockA{
\textsuperscript{1}\textit{
School of Computer Science, Shanghai Key Laboratory of Intelligent Information Processing, Fudan University, China}\\
\textsuperscript{2}\textit{
Academy for Engineering and Technology, Fudan University, China}\\
\textsuperscript{3}\textit{
Department of Ophthalmology, Shanghai General Hospital, School of Medicine, Shanghai Jiao Tong University, China}\\
\textsuperscript{4}\textit{Shanghai Eye Diseases Prevention \& Treatment Center, Shanghai Eye Hospital, Shanghai, China}\\
\texttt{\{jlhou18,yjzhang,fengrui\}@fudan.edu.cn}
}
}

\maketitle

\begin{abstract}
Automatic diabetic retinopathy (DR) grading based on fundus photography has been widely explored to benefit the routine screening and early treatment. Existing researches generally focus on single-field fundus images, which have limited field of view for precise eye examinations. In clinical applications, ophthalmologists adopt two-field fundus photography as the dominating tool, where the information from each field (i.e., macula-centric and optic disc-centric) is highly correlated and complementary, and benefits comprehensive decisions. However, automatic DR grading based on two-field fundus photography remains a challenging task due to the lack of publicly available datasets and effective fusion strategies. In this work, we first construct a new benchmark dataset (DRTiD) for DR grading, consisting of 3,100 two-field fundus images. To the best of our knowledge, it is the largest public DR dataset with diverse and high-quality two-field images. 
Then, we propose a novel DR grading approach, namely Cross-Field Transformer (CrossFiT), to capture the correspondence between two fields as well as the long-range spatial correlations within each field. Considering the inherent two-field geometric constraints, we particularly define aligned position embeddings to preserve relative consistent position in fundus. Besides, we perform masked cross-field attention during interaction to filter the noisy relations between fields. Extensive experiments on our DRTiD dataset and a public DeepDRiD dataset demonstrate the effectiveness of our CrossFiT network. The new dataset and the source code of CrossFiT will be publicly available at \url{https://github.com/FDU-VTS/DRTiD}.
\end{abstract}


\section{Introduction}

\begin{figure}[t]
\includegraphics[width=\columnwidth]{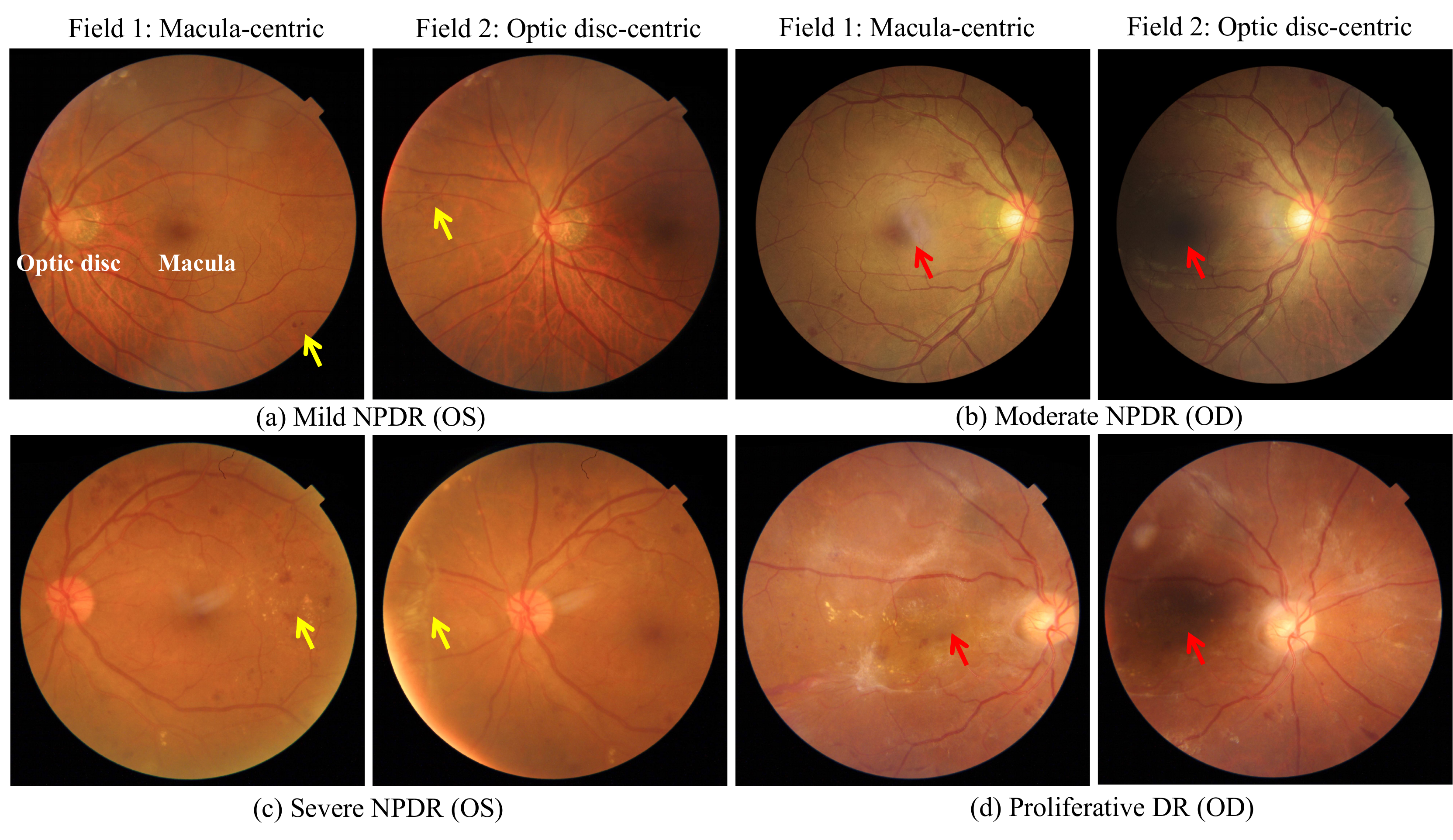}
\caption{
Examples of two-field fundus images from our DRTiD dataset.
(a) and (c) show the Oculus Sinister (OS) with mild and severe NPDR. 
(b) and (d) show the Oculus Dexter (OD) with moderate NPDR and PDR. 
} 
\label{intro}
\end{figure}

Diabetic retinopathy (DR) is a leading cause of visual impairment and blindness in working-age population worldwide \cite{zheng2012worldwide}.  
According to the International Clinical Diabetic Retinopathy Scale \cite{wilkinson2003proposed}, the disease severity of DR can be divided into five stages, including No Apparent Retinopathy, Mild Nonproliferative DR (NPDR), Moderate NPDR, Severe NPDR, and Proliferative DR (PDR). 

Digital fundus photography is a major tool in retinal screening, which usually involves various fields of view.
For instance, single‑field fundus photography is a convenient and widely-used technique, which takes the midpoint between the macula and optic disc as the center of shooting field. However, a great limitation is that the covered view of single field is insufficient to detect peripheral lesions \cite{single-field}. Thus, it cannot support a comprehensive eye examination.
Instead of single-field, two-field fundus photography has become the dominating method for DR screening at the community level \cite{comparison}. As illustrated in Fig. \ref{intro}, two fields take the macula and optic disc as the center of shooting field, respectively. 
There are several significant advantages of two-field images compared to single-field images. (1) Two-field images expand the coverage of examined retina areas, so that some periphery lesions can be comprehensively observed (e.g., yellow arrows in Fig. \ref{intro}); (2) There is a natural complementarity between two-field images. When the fundus regions in one image are covered in shadow or artifact, ophthalmologists can refer to the corresponding area in the other image (e.g., red arrows in Fig. \ref{intro}). 

Deep learning approaches have demonstrated significant improvement in DR screening. The existing research works primarily concentrate on single-field fundus photography.
However, the research on two-field fundus photography for automatic DR diagnosis is still largely unexplored. The reason is two-fold: (1) The publicly available datasets are extremely scarce. (2) Most of the two-field approaches mainly concentrate on simple fusion strategies. As two-field fundus images have natural geometric correspondences, the key challenge is how to integrate the information from two fields under the guidance of domain knowledge.

In this paper, we first construct a new benchmark dataset for DR grading, namely Diabetic Retinopathy Two-field image Dataset (DRTiD). It consists of 3,100 two-field fundus images from 1,550 examined eyes. This dataset is very significant to the current community in view of its four properties, i.e., rich diversity, high quality, large scale, and availability. 
Then, we propose a novel Cross-Field Transformer (CrossFiT) for DR grading on two-field fundus images. 
Specifically, our CrossFiT can capture both correspondence between two fields and long-range spatial correlations within each field. According to the standard two-field screening mechanism, we define aligned position embeddings to retain geometric constraints between two fields. Moreover, we introduce masked cross-field attention to prevent the noisy information propagation caused by redundant areas in two images.
We demonstrate the effectiveness of CrossFiT on both our DRTiD dataset and a public DeepDRID dataset. In summary, our main contributions are the following three aspects:
\begin{itemize}
\item We construct the largest public dataset DRTiD with diverse and high-quality two-field images for DR grading. 
This new benchmark will be beneficial to further research on DR diagnosis with two-field fundus images.

\item We propose a novel Cross-Field Transformer (CrossFiT), which effectively exploits the two-field correspondences and boosts DR grading performance. 
It learns customized two-field interaction with aligned position embeddings and masked cross-field attention mechanism. 

\item Comprehensive experiments and analyses on DRTiD and DeepDRiD datasets demonstrate the superiority of our CrossFiT for DR grading with two-field fundus images.
\end{itemize}

\section{Related Work}

\subsection{Diabetic Retinopathy Grading}
For DR grading with two-field fundus images, the early work \cite{autodetsystem} identified DR by counting specific lesions and aggregating them from each field.
Recently, a simple yet effective decision-level fusion strategy was widely adopted in two-field DR grading \cite{access,naturedr}. It followed the same training process as single-field methods but accepted the severer grade between two images as the final result when testing.
However, directly assigning the eye-level ground-truth label to each field for separate training would cause significant label noise.
To address this issue, Fang et al. \cite{slr} refined training labels for each image by a Stochastic Label Refinery method and averaged the two predicted scores in the test phase.
Nevertheless, these existing attempts only built the two-field connections at the last decision level, while leaving the underlying feature-level correlations greatly unexploited.

\subsection{Multi-view Medical Image Analysis}
Multi-view classification approaches have been widely explored in medical image analysis.
For example, Qian et al. \cite{qian2021two} proposed a two-stream binocular network to capture the correlations between left and right eyes with the contrastive grading loss.
For chest X-ray, Rubin et al. \cite{rubin2018large} designed a DualNet for thoracic disease prediction by aggregating the features from frontal and lateral images.
Hashir et al. \cite{hashir2020quantifying} compared several multi-view methods to merge the posteroanterior and lateral views of X-ray images for predicting radiological results.
These methods focus on learning multi-view correlations at the feature level. 
Different from other types of medical data, two-field fundus photography has its explicit correspondences caused by screening mechanism, which helps to design stronger customized two-field interaction mechanism.

\section{A New Benchmark Dataset}


There are only a few two-field DR image datasets in the current research community. Sharath et al. \cite{autodetsystem} collected 1,344 two-field images, yet they were not released for public research. 
DeepDRiD \cite{deepdrid} was the only publicly available dataset with 2,000 two-field fundus images. 
However, the image scale and quality of the existing datasets remain inadequate for the profound study of two-field DR grading.
In this work, we construct a new benchmark dataset, namely Diabetic Retinopathy Two-field image Dataset (DRTiD).
It provides 3,100 two-field fundus images with rich diversity and high quality, taken from multiple practical screening scenarios.
This large-scale dataset is very significant as it makes up an important shortfall in the current research community and catalyzes future research in two-field DR diagnosis.

We select images from Shanghai Diabetic Eye Study (SDES) between 2015 and 2017.
All the images are captured with digital 45° to 55° non-mydriatic retinal cameras.
The image resolutions range from 1,444$\times$1,444 to 3,058$\times$3,000 pixels. 
Personal information appearing on fundus images and file names is removed to protect privacy.
Data cleaning is conducted to guarantee the high image quality and correct fields of view. The qualified images should be of gradable quality in clarity, illumination, and artifacts. Besides, each pair contains macula-centric and optic disc-centric fundus images of an examined eye.

The full set of 3,100 images is annotated independently by a team of three experienced ophthalmologists. Annotations include the eye-level DR \& DME (Diabetic Macular Edema, a complication associated with DR) severity grade, and the localization of optic disc and macula. 
The images with divergent annotations are further confirmed by an expert ophthalmologist with clinical experience of more than ten years. 
According to the International Clinical Diabetic Retinopathy Scale \cite{wilkinson2003proposed}, all images are graded into five groups, i.e., No-DR, Mild NPDR, Moderate DR, Severe NPDR, and PDR. Our dataset contains 747/140/406/199/58 pairs of two-field fundus images for each grade, respectively. 
We divide the labeled data into the training set and testing set, which are comprised of 2,000 and 1,100 two-field fundus images.
Both sets maintain an appropriate mixture of disease stratification and sample diversity. Some examples of two-field fundus images from our DRTiD dataset are shown in Fig. \ref{intro}.

\begin{figure*}[t]
\includegraphics[width=\textwidth]{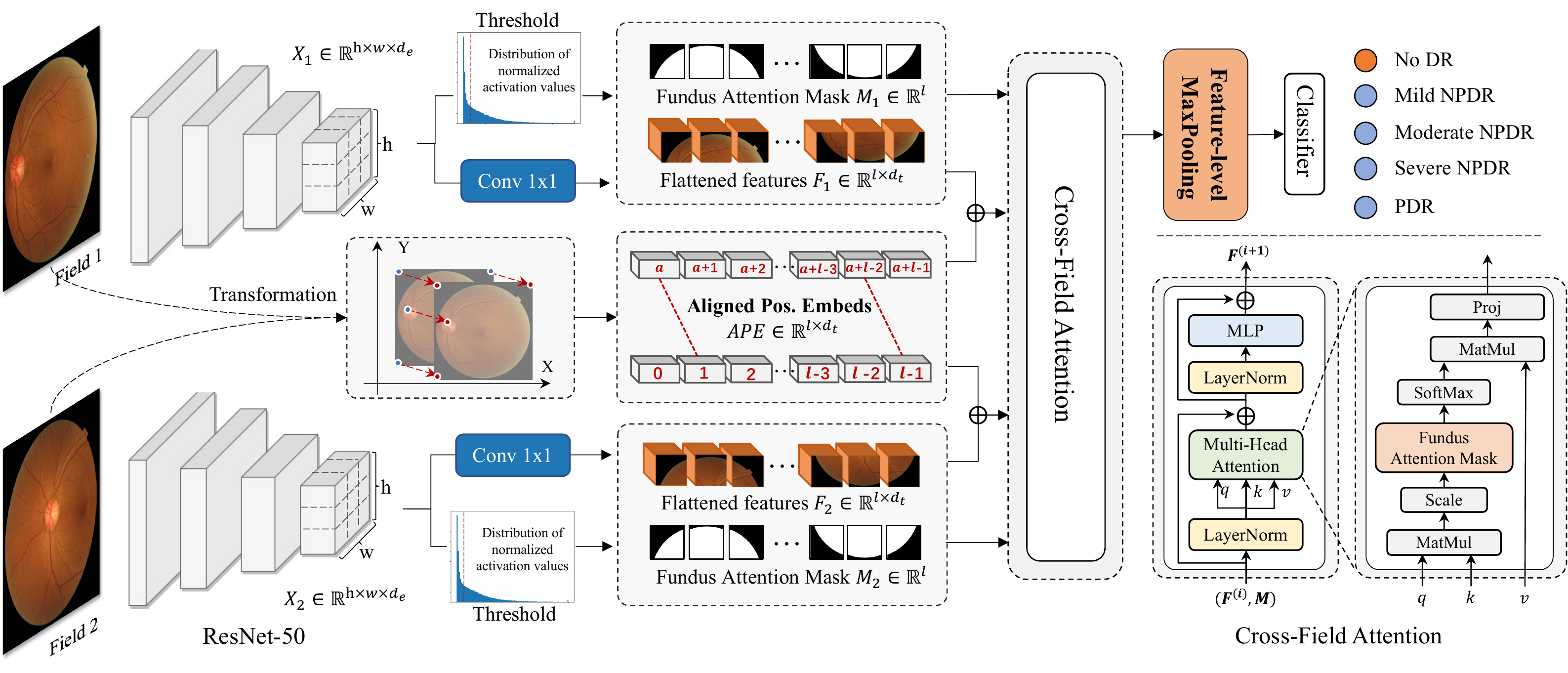}
\caption{The overall structure of our Cross-Field Transformer. It consists of the encoder (ResNet-50), Cross-Field Attention, and feature-level maxpooling operation. The flattened features, aligned position embeddings, and fundus attention masks are taken as the inputs to Cross-Field Attention module.}
\label{framework}
\end{figure*}

\section{Methodology}

\subsection{Problem Formulation}
Given two-field fundus images $\mathbf{I}_1, \mathbf{I}_2\in \mathbbm{R}^{H\times W\times 3}$ of an eye, we aim to train a deep network $f_\theta$ to predict the eye-level DR grade $\mathbf{\hat{Y}}=f_\theta(\mathbf{I}_1,\mathbf{I}_2)$ over $C$ categories. Specifically, the network $f_\theta = \{E_\theta, S_\theta\}$ is composed of the encoder $E_\theta$ and the classifier $S_\theta$, parameterized by $\theta$. A standard cross-entropy loss $\mathcal{L}$ is used to minimize the difference between the ground truth $\mathbf{Y}$ and prediction $\mathbf{\hat{Y}}$.

\subsubsection{Single-field methods}

Single-field methods predict the DR grade based on each field separately, i.e.,  $\mathbf{\hat{Y}}_1=f_\theta(\mathbf{I}_1),~ \mathbf{\hat{Y}}_2=f_\theta(\mathbf{I}_2)$. 
The loss function is given by $\mathcal{L}(\mathbf{Y},\mathbf{\hat{Y}}_1)+\mathcal{L}(\mathbf{Y},\mathbf{\hat{Y}}_2)$, where the eye-level grade $\mathbf{Y}$ is assigned to each field as pseudo labels. 
However, there are two drawbacks of single-field methods. (1) They miss the two-field relationship and cannot obtain eye-level predictions. (2) The noise in pseudo labels affects the grading performance.

\subsubsection{Two-field methods}
The main purpose of two-field methods is to model relationships and fuse information from two images effectively. 
\textbf{Decision-level fusion} integrates the predictions of two fields into the final prediction. 
Specifically, the network predicts two images separately, and then averages the probabilities $\mathbf{P}=\text{avg}(\mathbf{P}_1,\mathbf{P}_2)$, or selects the severer grade $\mathbf{\hat{Y}}=\max\{\mathbf{\hat{Y}}_1,\mathbf{\hat{Y}}_2\}$. 
Nevertheless, simply fusing predictions fails to capture the spatial and semantic relationship between two fields. 
\textbf{Feature-level fusion} fuses the deep features of each field $\mathbf{X}_1=E_\theta(\mathbf{I}_1)$ and $\mathbf{X}_2=E_\theta(\mathbf{I}_2)$ to form a unified feature representation. Then, the DR grade is predicted based on both features $\mathbf{\hat{Y}}=S_\theta(\mathbf{X}_1,\mathbf{X}_2)$. Common strategies include feature average pooling and concatenation. Unlike these simple operations that require no learnable parameters, we introduce an additional fusion module to learn customized spatial and semantic correspondences and boost DR grading performance.

\subsection{Cross-Field Transformer for DR Grading}

As depicted in Fig. \ref{framework}, our Cross-Field Transformer takes two-field fundus images as the input and outputs the DR severity grade in an end-to-end manner. Firstly, we adopt ResNet-50 as the encoder to produce feature maps $\mathbf{X}_1, \mathbf{X}_2 \in \mathbbm{R}^{h\times w\times d_e}$ of each field, where $h, w, d_e$ represent the height, width, and channels, respectively. Then, the feature maps are flattened and linearly projected into the sequence vectors with $d_t$ dimension, i.e., $\mathbf{F}_1, \mathbf{F}_2 \in \mathbbm{R}^{l\times d_t}$, where $l=h\times w$ is the sequence length.
We particularly introduce the aligned position embeddings $\mathbf{APE}\in\mathbbm{R}^{l\times d_t}$ to retain the corresponding geometric relation between two fields. In addition, fundus attention masks $\mathbf{M}_1, \mathbf{M}_2=\{0,1\}\in\mathbbm{R}^{l}$ are generated by applying a threshold on the averaged activation values of the feature maps to avoid the noisy effect caused by redundant regions. The feature sequences, added with position embeddings, serve as inputs for the following Cross-Field Attention module.
The module simultaneously models long-range spatial correlations within each field and constructs correspondence between two fields. 
Finally, we employ the maxpooling operation on two global features, which is first found more effective than the commonly-used average and concatenation operations for two-field DR grading. A classifier predicts the DR grade based on the fused features. 

\subsubsection{Cross-Field Attention}
To encode the inter-field and intra-field information, two sequences are concatenated as the entire input, i.e., $\mathbf{F}=\text{concat}[\mathbf{F}_1; \mathbf{F}_2]\in\mathbbm{R}^{2l\times d_t}$.
As shown in Fig. \ref{framework}, the Cross-Field Attention module is a stack of layers of multi-head self-attention (MSA) and multi-layer perception (MLP) blocks. LayerNorm is applied before every block, and residual connections after every block. The MLP contains two layers with a GELU non-linearity.

In the multi-head self-attention, for the $n$-th head, linear projections $\mathbf{W}_n^Q, \mathbf{W}_n^K, \mathbf{W}_n^V \in \mathbbm{R}^{d_t\times d_t/N}$ are adopted to obtain the queries, keys and values, denoted as $\mathbf{Q}_n=\mathbf{F}\mathbf{W}_n^Q, \mathbf{K}_n=\mathbf{F}\mathbf{W}_n^K, \mathbf{V}_n=\mathbf{F}\mathbf{W}_n^V \in \mathbbm{R}^{2l\times d_t/N}$,
where $n=1,2,...,N$ is the number of attention heads. The attention weight $\mathbf{A}_n=\mathbbm{R}^{2l\times 2l}$ models both the intra-field interdependencies and inter-field correspondences between different positions. It is calculated as:
\begin{equation}
    \mathbf{A}_n=\mathrm{softmax}\left(\mathbf{Q}_n \mathbf{K}_n^\top/\sqrt{d_t/N}\right),
\label{eq: attention}
\end{equation}
where $\sqrt{d_t/N}$ is a scaling factor. The output of the head $\mathbf{H}_n\in\mathbbm{R}^{2l\times d_t/N}$ is computed by matrix multiplication of the weight $\mathbf{A}_n$ and values $\mathbf{V}_n$, i.e., $\mathbf{H}_n=\mathbf{A}_n\mathbf{V}_n$.
Then, all single head outputs $\{\mathbf{H}_n\}_{n=1}^N$ are concatenated along the channel dimension to obtain the multi-head output $\mathbf{H}\in\mathbbm{R}^{2l\times d_t}$ through a projection matrix $\mathbf{W}^O\in\mathbbm{R}^{d_t\times d_t}$,
\begin{equation}
    \mathbf{H}=\textrm{concat}[\mathbf{H}_1;\mathbf{H}_2;...;\mathbf{H}_N]\mathbf{W}^O.
\end{equation}


\subsubsection{Aligned Position Embeddings}
Position embeddings provide important positional information of each patch to the Transformer network. 
Given a grid of patches in $X$ and $Y$ dimensions, we apply sine and cosine functions \cite{vaswani2017attention} to produce 2-D positional embeddings for each field. The $X$-embedding and $Y$-embedding with size $d_t/2$ are encoded separately by the following equation:
\begin{equation}
\begin{aligned}
    \mathbf{PE}_{(pos,2i)}&=sin(pos/10000^{2i/(d_t/2)}),\\
    \mathbf{PE}_{(pos,2i+2)}&=cos(pos/10000^{2i/(d_t/2)}),
\end{aligned}
\label{eq: cos}
\end{equation}
where $pos$ is the position and $i$ is the dimension. The two embeddings are concatenated to obtain the 2-D positional embeddings $\mathbf{PE}\in\mathbbm{R}^{l\times d_t}$.

A common practice is that two fields share the regular grid of patches and generate the same position embeddings, i.e., $PE_1=\mathbf{PE}, PE_2=\mathbf{PE}$.
However, the pair of individual grids misses the global geometric constraints between two-field images. 
We hereby integrate two grids into a unified coordinate and produce aligned position embeddings. 
For simplicity, if we treat $\mathbf{I}_2$ as the moving image, the aligned grid of the fixed image $\mathbf{I}_1$ can be obtained by transforming $\mathbf{I}_2$ to be aligned with $\mathbf{I}_1$. 
According to the fundus screening mechanism, two fields can be aligned approximately by a simple transformation matrix $\mathcal{T}_\alpha$ with translation parameters. 
We model the customized geometric constraints for each pair by utilizing the optic disc as the visual cue. 

Formally, let $\mathbf{G}=\{\mathbf{G}_i\}$ of pixels $\mathbf{G}_i=(\mathbf{x}_i,\mathbf{y}_i)$ be the regular grid of the image $\mathbf{I}_2$. We have the relative coordinates of the optic disc in each field, denoted as $(\mathbf{x}^1_{op},\mathbf{y}^1_{op}), (\mathbf{x}^2_{op},\mathbf{y}^2_{op})$. The point-wise translation to obtain the aligned grid $\mathbf{AG}$ of $\mathbf{I}_1$ can be written as:
\begin{equation}
\begin{pmatrix}
\mathbf{\tilde{x}}_i\\
\mathbf{\tilde{y}}_i
\end{pmatrix} = \mathcal{T}_\alpha(\mathbf{G}_i)=
\begin{bmatrix}
1 & 0 & 2\times(\mathbf{x}^2_{op}-\mathbf{x}^1_{op})\\
0 & 1 & 2\times(\mathbf{y}^2_{op}-\mathbf{y}^1_{op})
\end{bmatrix}
\begin{pmatrix}
\mathbf{x}_i\\
\mathbf{y}_i\\
1
\end{pmatrix},
\end{equation}
where $(\mathbf{x}_i,\mathbf{y}_i)$ are the original coordinates in $\mathbf{G}$, and $(\mathbf{\tilde{x}}_i,\mathbf{\tilde{y}}_i)$ are the transformed coordinates in $\mathbf{AG}$. 

We use height and width normalized coordinates, such that $-1\leq \mathbf{x}_i,\mathbf{y}_i\leq 1$. 
Afterwards, we down-sample the image-level grids to the feature-level shape $h\times w\times 2$ by the bilinear interpolation algorithm and denormalize them by height and width. With the Eq. \ref{eq: cos}, the aligned grids $\mathbf{AG}$ of image $\mathbf{I}_1$ and $\mathbf{G}$ of image $\mathbf{I}_2$ are calculated to obtain the aligned position embeddings, i.e., $PE_1=\mathbf{APE}, PE_2=\mathbf{PE}$.

\begin{table*}[t]
\begin{center}
\caption{Comparison results of our proposed method and existing single-field and two-field methods on our DRTiD dataset. (Unit: \%)}
\label{table:comparison}
\begin{tabular}{llccccccccc}
\toprule
\multirow{2}{*}{Category} & \multirow{2}{*}{Method} & \multirow{2}{*}{Backbone} & 
\multirow{2}{*}{Kappa} & \multirow{2}{*}{Acc} & \multirow{2}{*}{AUC} & \multicolumn{5}{c}{AUC} 
\\
\cmidrule(lr){7-11}
& & & & & & 0 & 1 & 2 & 3 & 4\\
\midrule
Single-field
 & Zoom-in-Net \cite{zoomin} & Inception-ResNet & 77.33 & 71.00 & 80.46 & 91.51 & 62.96 & 80.36 & 91.27 & 75.92\\
 & Lesion-based CL \cite{huang2021lesion} (Regression)& ResNet-50 & 79.15 & 65.00 & - & -& -& -& -& -\\
 & Both Fields & ResNet-50 & 78.06 & 70.91 & 85.96 & 91.81 & 71.17 & 80.71 & 94.02 & 91.76\\
 \cmidrule{2-11}
 & Macula-centric Field & ResNet-50 & 80.47 & 73.09 & 86.69 & 92.47 & 72.27 & 82.06 & 94.25 & 91.37\\
 & Optic dist-centric Field & ResNet-50 & 77.87 & 70.91 & 84.73 & 91.85 & 67.64 & 79.63 & 94.16 & 89.58\\
\midrule
Two-field
& Binocular Network \cite{qian2021two} & ResNet-50 & 77.85 & 66.18 & 82.86 & 90.70 & 66.42 & 78.33 & 92.25 & 86.23\\
& Cross-view Transformer \cite{tulder2021multi} & ResNet-50 & 80.54 & 69.45 & 87.74 & 92.22 & 73.61 & 81.94 & 94.60 & \textbf{95.70}\\
& Prediction Average \cite{2020DeepTMI} & ResNet-50 & 80.49 & 73.82 & 86.07 & 94.28 & \textbf{75.24} & 83.76 & 94.49 & 81.88\\
& DeepDR \cite{naturedr} & ResNet-101 & 81.60 & 72.73 & 87.37 & 94.15 & 68.82 & 84.48 & 95.22 & 93.49\\
& Prediction Max \cite{access} & ResNet-50 & 82.53 & 73.09 & 87.66 & 93.65 & 71.85 & 84.13 & 94.41 & 93.56\\
& Feature Concatenation \cite{rubin2018large} & ResNet-50 & 82.71 & 74.36 & 88.46 & 94.55 & 73.35 & 85.58 & 94.60 & 93.47\\
& Feature Average \cite{hashir2020quantifying} & ResNet-50 & 82.73 & 75.27 & 87.92 & 94.60 & 73.00 & 84.77 & 93.57 & 92.92\\
\cmidrule{2-11}
& CrossFiT & ResNet-50 & \textbf{84.21} & \textbf{75.64}  & \textbf{88.50} & \textbf{94.70} & 70.24 & \textbf{85.88} & \textbf{95.51} & 95.47 \\
\bottomrule
\end{tabular}
\end{center}
\end{table*}

\subsubsection{Fundus Attention Mask}
A fundus image contains redundant regions, such as the black background in the corner, and occasional small areas of shadow or artifacts in the fundus region. Noise caused by these regions can be involved in the information propagation procedure. 
To address this issue, we particularly introduce a fundus attention mask (FAM) into the multi-head attention module, which helps to filter the noisy relations.
Given the feature maps $\mathbf{X}_1, \mathbf{X}_2\in\mathbbm{R}^{h\times w\times d_e}$, we average the feature maps in channel and obtain activation maps $\mathbf{\bar{X}}_1, \mathbf{\bar{X}}_2\in\mathbbm{R}^{h\times w}$. Then, the values of activation maps are normalized into the interval $[0,1]$. We set a threshold $p\in[0,1]$ to split fundus and redundant areas.
\begin{equation}
    \mathbf{M}(i,j)=\left\{
    \begin{aligned}
        &1, &\textrm{if}~\mathbf{\bar{X}}(i,j)\geq p;\\
        &0, &\textrm{otherwise}.
    \end{aligned}
    \right.
\end{equation}
We can obtain flattened fundus attention masks $\mathbf{M}_1, \mathbf{M}_2\in\{0,1\}^{\mathbbm{R}^{l}}$, where $l=h\times w$. In the multi-head attention module, we use the masks inside of scaled dot-product attention by masking out (setting to $-\infty$) the redundant values before softmax. 
The Eq. \ref{eq: attention} can be replaced as:
\begin{equation}
    \mathbf{A}_n=\mathrm{softmax}\left(\mathbf{M}(\mathbf{Q}_n\mathbf{K}_n^\top/\sqrt{d_t/N})\right).
\end{equation}
According to our observation, the choice of threshold $p$ is dependent to data samples. The optimal thresholds are 0.06 and 0.03 for DRTiD and DeepDRiD datasets, respectively.

\section{Experiments}

\subsection{Datasets and Evaluation Metrics}
EyePACS dataset \cite{kaggle} contains 35,126/10,906/42,670 images for training/validation/testing. We use the training images to pre-train the encoder in CrossFiT.
DeepDRiD dataset \cite{deepdrid} includes 1,000 pairs of two-field images, which are divided into 600/200/200 for train/public test/private test set. 
Our constructed DRTiD dataset provides 3,100 two-field fundus images from 1,550 eyes, which are split into 2,000/1,100 training/testing images. 
We adopt Quadratic Weighted Kappa \cite{kaggle}, Accuracy, and Macro AUC for overall comparison, and AUC for each grade as evaluation metrics. We also report statistical analysis based on the independent two-sample t-test.

\subsection{Implementation Details}
We use ResNet-50 \cite{resnet} pre-trained on EyePACS dataset as the backbone network. All images are resized to 512$\times$512. Data augmentations include random crop, rotation, and color jittering. We adopt SGD with a momentum of 0.9, a weight decay of 1e-5, and a batch size of 64. The networks are trained for 100 epochs with the initial learning rate 1e-3.
We employ a 3-layer cross-field attention with $d_t$=1024.
Our method is implemented in PyTorch and runs on Tesla V100 GPUs.

\subsection{Results on Our DRTiD Dataset}

\subsubsection{Comparison with single-field methods}
We first compare the CrossFiT with several representative single-field methods. 
The Zoom-in-Net \cite{zoomin} and Lesion-based CL \cite{huang2021lesion} networks are re-implemented on our dataset. 
We also train a ResNet-50 model on both fields as the baseline in our work.
As can be seen from the first three rows in Table \ref{table:comparison}, the three single-field methods obtain the Kappa scores of 77.33\%, 79.15\%, and 78.06\%, respectively. Despite the significant improvement on single-field DR grading, these single-field methods still achieve inferior performance on two-field fundus images. 
Moreover, we report the diagnostic results of training and testing ResNet-50 models using only macula-centric or optic disc-centric field. The 4th and 5th rows in Table \ref{table:comparison} show that the two methods obtain similar performance, which demonstrates that neither field achieves a dominant position, but each field contributes almost equally to two-field DR grading.

\subsubsection{Comparison with two-field methods}
We next compare the CrossFiT with other existing cross-view methods, including commonly-used feature-level and decision-level fusion strategies.
As can be seen in the two-field category in Table \ref{table:comparison}, our proposed CrossFiT network clearly outperforms other two-field methods on the three metrics. It achieves 84.21\% Kappa, 75.64\% Acc, and 88.50\% Macro AUC, which demonstrates the superior two-field DR grading results. Besides, the CrossFiT reaches high AUC scores on each grade, especially on the 0, 3, and 4 grades with about 95\% AUC. The results show the effectiveness of our CrossFiT to capture the correlations between two-field images and achieve better performance.

\subsection{Ablation Study}

The ablation study on each module in our CrossFiT network is shown in Table \ref{table: DRTiD, ablation hidden size}. Compared to the single-field ResNet-50, the group of two-field models achieve significant improvements. As a feature-level fusion operation, maxpooling (83.08\% Kappa) is demonstrated more effective for two-field images than the commonly-used concatenation (82.71\% Kappa) and average (82.73\% Kappa) fusion operations. By employing the Cross-Field Attention (CFA) module for information propagation, the Kappa score increases to 83.81\%. Furthermore, the APE and FAM also greatly benefit the DR grading performance. 
In the following paragraphs, we will analyse the effectiveness of APE and FAM in detail.

\begin{table}[t]
\begin{center}
\caption{Ablation study of our CrossFiT on DRTiD dataset. The numbers in square brackets represent the 95\% confidence interval.}
\label{table: DRTiD, ablation hidden size}
\begin{tabular}{lccccc}
\toprule
Method & Maxpool & CFA & APE & FAM & Kappa \\
\midrule
 ResNet-50 & & & & & 78.06 [75.48, 80.51] \\
\midrule
 & $\checkmark$ & & & & 83.08 [80.04, 85.90] \\
 & $\checkmark$ & $\checkmark$ & & & 83.81 [80.96, 86.65] \\
 & $\checkmark$ & $\checkmark$ & $\checkmark$ & & 83.97 [81.01, 86.75] \\
\midrule
 CrossFiT & $\checkmark$ & $\checkmark$ & $\checkmark$ & $\checkmark$ & 84.21 [81.47, 87.00] \\
\bottomrule
\end{tabular}
\end{center}
\end{table}

\subsubsection{Aligned position embeddings}

\begin{table}[t]
\begin{center}
\caption{The comparison of different position embeddings of CrossFiT.}
\label{table: DRTiD: ablation APE}
\begin{tabular}{p{80pt}p{100pt}<{\centering}}
\toprule
Pos. Embeds & Kappa \\
\midrule
w/o Pos. Embeds & 82.99 [80.04, 85.85] \\
learnable Pos. Embeds &  83.22 [80.33, 86.04]\\
regular Pos. Embeds & 83.84 [80.98, 86.67]\\
aligned Pos. Embeds & 84.21 [81.47, 87.00]\\
\bottomrule
\end{tabular}
\end{center}
\end{table}

The positional information is important for two-field DR grading. As can be seen from Table \ref{table: DRTiD: ablation APE}, we compare three different ways of positional embedding, including learnable Pos. Embeds, regular (cosine) Pos. Embeds, and our proposed aligned Pos. Embeds. The results show that both learnable Pos. Embeds and regular Pos. Embeds surpass w/o Pos. Embeds. However, they lack the corresponding spatial relationships between two-field images. In comparison, the aligned Pos. Embeds provide global geometric prior correlations, outperforming w/o Pos. Embeds by 1.22\% Kappa score.

\subsubsection{Threshold $p$ in the Fundus Attention Mask}


\begin{table}[t]
\begin{center}
\caption{Results of different threshold $p$ in FAM.}
\label{table: DRTiD: ablation FAM}
\begin{tabular}{lccccccccc}
\toprule
$p$ & 0.02 & 0.04 & 0.05 & 0.06 & 0.07 & 0.08 & 0.10 \\
\midrule
Kappa & 83.55 & 83.44 & 83.78 &  84.21 & 83.97 & 83.74 & 83.43\\
\bottomrule
\end{tabular}
\end{center}
\end{table}


We investigate the choice of threshold $p$ to generate fundus attention masks.
The comparison results of different thresholds are presented in Table \ref{table: DRTiD: ablation FAM}, where $p=$0.06 is chosen as the optimal threshold. When $p$=0, all the pixels are considered to influence each other in the self-attention. If $p$ falls in (0, 0.6), the masks would contain irregular foreground and irrelevant region. 
The masks obtained by the optimal activation threshold $p$=0.06 can highlight fundus regions, which exclude redundant black borders and partial low-quality regions precisely. 
With the increase of $p$, it may cause information loss.

\subsection{Results on the DeepDRiD Challenge Leaderboard}
Table \ref{table: deepdrid} shows the results of our CrossFit network and other challenge participation methods on the test sets of DeepDRiD.
Team 1 \cite{slr} adopted an ensemble model of three EfficientNets (i.e., b5, b3, b1) and achieved 93.03\% and 92.15\% Kappa scores, which ranked first in the challenge. Team 3 \cite{deepdridteam3} utilized extra supervisions from 15 additional public datasets and reached 92.32\% and 90.97\% Kappa.
In comparison, our CrossFiT network is only pre-trained on EyePACS dataset and then fine-tuned on DeepDRiD dataset. When testing, we adopt test time augmentation technique \cite{DBLP:journals/corr/abs-2011-11156} to boost the generalization ability of our model. 
It is observed that the single CrossFiT achieves the best results of 93.33\% and 93.07\% Kappa scores on public test and private test sets, respectively. Our method is higher than the top-ranked result by Team 1 on the leaderboard, with 0.30\% and 0.92\% improvements.

\begin{table}[t]
\begin{center}
\caption{Comparison results on the DeepDRiD leaderboard. (Unit:\%)}
\label{table: deepdrid}
\resizebox{1\columnwidth}{!}{
\begin{tabular}{lccc}
\hline\noalign{\smallskip}
Method & Backbone & Public Test & Private Test\\
\noalign{\smallskip}
\hline
\noalign{\smallskip}
Team 1 \cite{slr}& \small{EfficientNet b5\&b3\&b1} & 93.03 & 92.15\\
Team 2 & Unpublished & 92.62 & 92.11\\
Team 3 \cite{deepdridteam3} & EfficientNet b4 & 92.32 & 90.97\\
Team 4 & Unpublished & 92.02 & 89.46\\
Team 5 & EfficientNet b7 & 90.88 & 88.90\\
\hline\noalign{\smallskip}
CrossFiT & ResNet-50 & \textbf{93.33} & \textbf{93.07}\\
\hline
\end{tabular}
}
\end{center}
\end{table}

\section{Conclusion}

In this paper, we conduct a profound research on the DR grading with two-field fundus photography. Firstly, we newly construct the largest dataset benchmark, Diabetic Retinopathy Two-field image Dataset (DRTiD), comprised of 3,100 high-quality and greatly diverse two-field fundus images. This dataset will be publicly available to encourage further studies on two-field fundus photography. Then, we propose a novel DR grading approach, namely Cross-field Transformer (CrossFiT), to exploit the correlation of two images and boost the performance of DR grading. 
Extensive experiments on the new DRTiD dataset and a public dataset have demonstrated the superiority of our CrossFiT approach. 

\bibliographystyle{IEEEtran}
\bibliography{IEEEfull}

\end{document}